\documentclass[runningheads,a4paper]{llncs}

\usepackage{amssymb}
\usepackage{graphicx}

\usepackage{url}
\urldef{\mailsa}\path|{alfred.hofmann, ursula.barth, ingrid.haas, frank.holzwarth,|
\urldef{\mailsb}\path|anna.kramer, leonie.kunz, christine.reiss, nicole.sator,|
\urldef{\mailsc}\path|erika.siebert-cole, peter.strasser, lncs}@springer.com|

\begin{document}

\mainmatter  

\title{The Tensor Memory Hypothesis}
\titlerunning{}

%
%
\author{Volker Tresp and  Yunpu Ma \\ Siemens AG and Ludwig Maximilian University of Munich}
\authorrunning{Tresp and Ma}

\institute{}

%
%

\toctitle{}
\tocauthor{}
\maketitle

\begin{abstract}








We discuss memory models which are based on tensor decompositions using  latent representations of entities and events. We show how episodic memory and semantic memory can be realized and  discuss how new memory traces can be generated from sensory input:  Existing memories are the basis for perception and new memories are generated via perception. We relate our mathematical approach to the hippocampal memory indexing theory.  We describe the first detailed mathematical models for  the complete processing pipeline from sensory input and its semantic decoding, i.e., perception,  to  the formation of episodic and semantic memories and their declarative semantic decodings.
Our main hypothesis is that perception includes an active  semantic decoding process, which relies on latent representations of entities and predicates, and that episodic and semantic memories depend on the same decoding process. We contribute to  the debate between  the leading memory consolidation theories, i.e., the standard consolidation theory (SCT) and the multiple trace theory (MTT). The latter is closely related to the complementary learning systems (CLS) framework. {In particular,  we show explicitly how episodic memory can teach the neocortex to form a semantic memory, which is a core issue in MTT and CLS.}


\end{abstract}

\section{Introduction}

It still is a great puzzle that the brain can easily deal with continuous high-dimensional signals, such as images and sounds, but at the same time relies on very discrete concepts like ``Jennifer Aniston''~\cite{quiroga2013brain}.  The latter faculty is most apparent in language, where we clearly make statements about discrete entities:
``Jennifer Aniston is an American actress, who became famous through the television sitcom \textit{Friends}''. We argue that a key concept might be representation learning, i.e., that each discrete entity is represented as a vector of real numbers.  Based on the latent representations,  different declarative memory functions can be implemented, in particular episodic memory and semantic memory. In this paper, we use the  framework of tensor decompositions to mathematically describe  the declarative nature of memory and perception.
 The effectiveness of tensor models for the realization of technical  memory functions
  is well established~\cite{nickel2012factorizing,nickel2016review}
 and here we explore their relevance in modelling human memories.
    Representation learning might also be the basis for perception: New memories are formed by mapping sensory inputs  to latent event representations which can be stored as episodic memories; these can then also be semantically decoded, using the tensor framework.  Thus sensory impressions can be  interpreted by the brain, become declarative and thus can verbally be described.   Perception,  in form of a semantic decoding of sensor input, as well as episodic and semantic memory,  all depend on the same latent representations, since the brain must know about entities and entity classes to understand and interpret new sensory inputs.
    { We describe the first detailed mathematical models for  the complete processing pipeline from sensory input and its semantic decoding, i.e., perception,  to  the formation of episodic and semantic memories and their declarative semantic decodings. }
    In contrast to a previous paper~\cite{LME-Nips2015}, here we relate our mathematical approach to the hippocampal memory indexing theory, which is one of the main theories for forming episodic memories~\cite{teyler1986hippocampal,teyler2007hippocampal}.  {In summary, our main hypothesis is that perception includes an active  semantic decoding process which relies on latent representations of entities and predicates and that episodic and semantic memories depend on the same decoding process.}

{    Eventually memory is consolidated from the hippocampal area to neocortex. In  the standard consolidation theory (SCT), episodic memory finds a representation in neocortex, essentially a copy of  the representations in hippocampus. In the multiple trace theory (MTT), episodic memory only is represented in the hippocampal area which is used to train semantic memory in neocortex. MTT is closely related to  the complementary learning systems (CLS) framework. We discuss the tensor memories in the context of both theories. } {In particular we show explicitly how episodic memory can teach the neocortex to form a semantic memory, which is a core issue in MTT and CLS.}

The paper is organized as follows. In the next section we describe the tensor memory models and in Section~\ref{sec:MM} we discuss different memory operations and illustrate how sensory inputs can generate new episodic memories. In Section~\ref{sec:cog} we discuss the potential relevance of the model for human perception and human memories.
In Section~\ref{sec:conso} we discuss memory consolidation.
Section~\ref{sec:concl2} contains our conclusions where we make the point that the brain might use the latent  representations in many additional functions, such as prediction, planning, and decision making.

%
%

%
%

\section{Tensor Memories}

Let $e_i$ be a symbol that represents entity $i$. We associate with  $e_i$ a latent vector $\mathbf{a}_{e_i}$.\footnote{An entity can, e.g.,  be a particular person or a location like ``Munich''.}
Similarly, we assume that a predicate $p$ is represented by a symbol $e_p$ with latent representation  $\mathbf{a}_{e_p}$.
We now propose that the probability that a triple statement\footnote{An example would be\textit{ (Jack, knows, Mary)}.} $(s, p, o)$ is part of semantic memory (``facts we know'') can be modelled as $P((s, p, o)) = \textit{sig}(\theta_{s, p, o})$, where $\textit{sig}(\cdot)$ is the  logistic function and $\theta_{s, p, o}$ is calculated as a function of the latent representations of the involved {predicate and}  entities,
\[
\theta_{s, p, o} = f^s(\mathbf{a}_{e_{s}},  \mathbf{a}_{e_{p}}, \mathbf{a}_{e_{o}})  .
\]
The function $f^s(\cdot)$ can be derived from  tensor factorization or can be realized as a  neural network. An overview of models can be found in~\cite{nickel2016review}.
To simplify the discussion
we focus on the Tucker model  with
\begin{equation}
\label{eq:sem}
  f^s(\mathbf{a}_{e_s}, \mathbf{a}_{e_p}, \mathbf{a}_{e_o}) =
 \sum_{r_1=1}^{\tilde r} \sum_{r_2=1}^{\tilde r} \sum_{r_3 =1}^{\tilde r}
 {a}_{e_s, r_1}   \;   {a}_{e_p, r_2}   \;  {a}_{e_o, r_3}  \;
 g^s(r_1, r_2, r_3)  .
\end{equation}
Here, $g^s(\cdot)$ is an element of the  core tensor.
The effectiveness of tensor models for the realization of technical semantic memory functions is well established~\cite{nickel2016review}.

For an episodic memory (``facts we remember'') we add a representation for time. Let  $e_t$ be the symbol for time instance $t$. Its latent representation then is   $\mathbf{a}_{e_t}$.
The probability of the observation of triple $(s, p, o)$ at time $t$ is
$P((s, p, o, t)) = \textit{sig}(\theta_{s, p, o, t})$ with
$
\theta_{s, p, o, t} = f^e(\mathbf{a}_{e_{s}},  \mathbf{a}_{e_{p}}, \mathbf{a}_{e_{o}}, \mathbf{a}_{e_{t}} )
$
and Tucker decomposition
\begin{equation}\label{eq:epis}
  f^e(\mathbf{a}_{e_s}, \mathbf{a}_{e_p}, \mathbf{a}_{e_o}, \mathbf{a}_{e_{t}}) =
 \sum_{r_1=1}^{\tilde r} \sum_{r_2=1}^{\tilde r} \sum_{r_3 =1}^{\tilde r}
 \sum_{r_4 =1}^{\tilde r}
 {a}_{e_s, r_1}   \;   {a}_{e_p, r_2}   \;  {a}_{e_o, r_3}  \; {a}_{e_t, r_4}  \;
 g^e(r_1, r_2, r_3, r_4)  .
\end{equation}

Whereas the episodic memory would be able to {retrieve} the fact that \textit{(Jack, diagnosed, Diabetes, Yesterday)}, the semantic memory would represent \textit{(Jack, diagnosed, Diabetes)}. Note that the tensor models can reconstruct known memories by assigning a high probability to  facts known to be true but they also assign high probabilities to facts which follow the patterns found in the memory systems; thus they realize a form of probabilistic inductive reasoning~\cite{nickel_three-way_2011}.  As an example,  consider that we know that Max lives in Munich. The probabilistic materialization that happens in the factorization should already predict that Max also lives in Bavaria and in Germany. Generalization from existing facts by probabilistic inductive reasoning is of particular importance in perception, where the predicted triple probability might serve as a prior for information extraction~\cite{dong_knowledge_2014,Baier2017ISWC}.
There is a certain danger in probabilistic materialization, since it might lead to  overgeneralizations, reaching  from  prejudice to false memories~\cite{roediger1995creating,loftus1996myth}. The effectiveness of tensor models for technical semantic memories is by now well established~\cite{nickel2015}.

\begin{figure}
  \centering
 \includegraphics[width=\columnwidth]{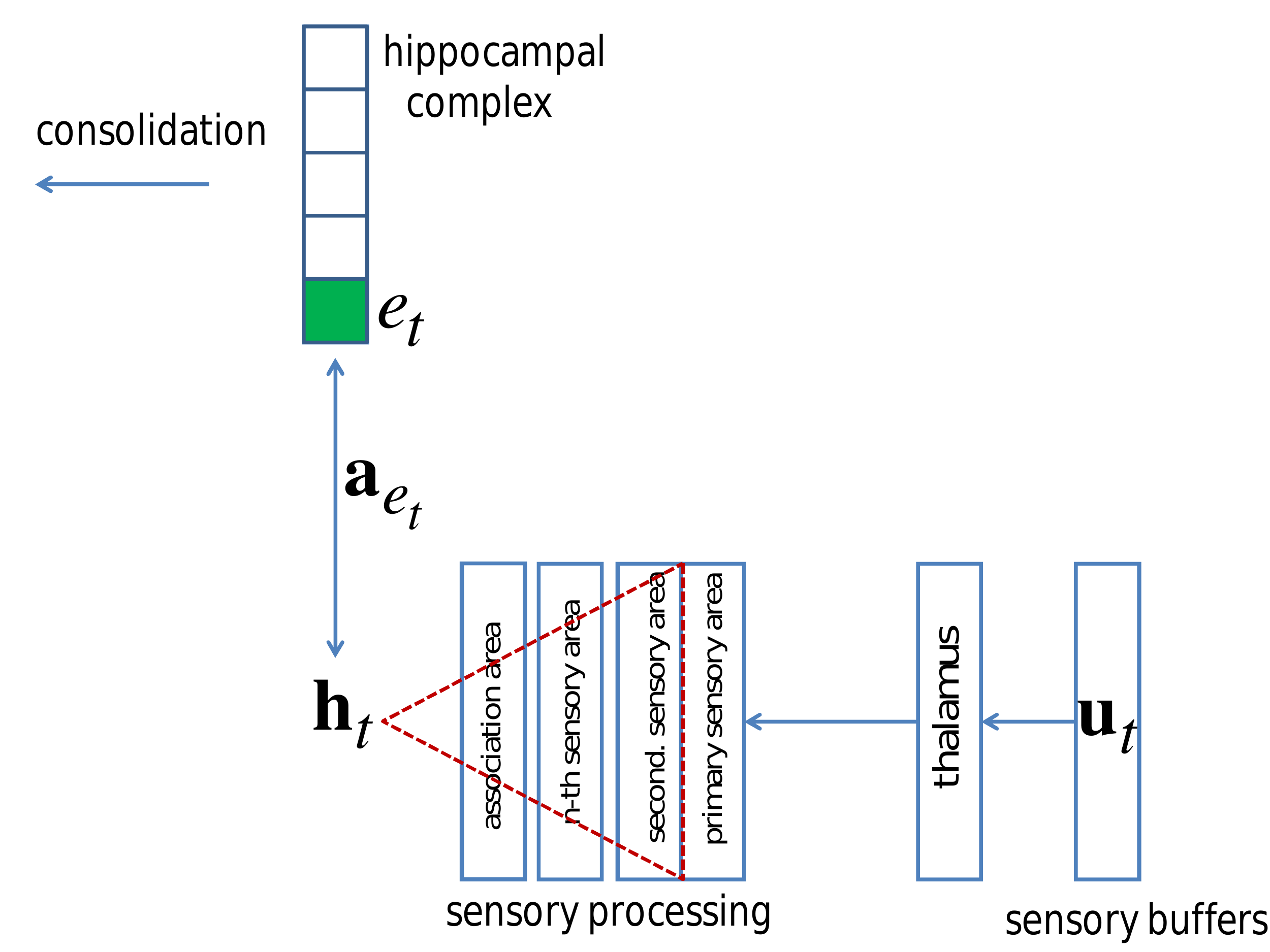}
  \caption{The figure sketches the hippocampal memory
indexing theory. Here, $\mathbf{u}_t$ represents the sensory input which is relayed through the thalamus (except olfaction), a part of the forebrain,  and then forms    the input to the different sensory processing paths.
Based on sensory processing, the latent representation for the time instance  $\mathbf{h}_t$ is generated.    As indicated, in the earlier  processing layers most nodes are activated, {while} at the higher layers, only a smaller number of scene-specific nodes are {active}. In the mathematical model, this mapping is described as
$\mathbf{h}_t = \mathbf{f}^M(\mathbf{u}_t)$.
  For sensory inputs to be stored as episodic memory ---because the event is e.g., significant, novel, or attached with emotion---     an index $e_t$ is formed in the hippocampus. $\mathbf{a}_{e_t}  = \mathbf{h}_t$ is  the sparse  weight pattern between the index and the sensory  layers which binds the sensory activation patterns with the index $e_t$.
  In recollecting  a memory  from a past  time $t'$, the corresponding index $e_{t'}$ is activated, which then reactivates an approximate $\mathbf{h}_{t'}$ via weight patterns $\mathbf{a}_{e_{t'}}$.
%
%
  }
  \label{fig:HMIT}
\end{figure}

\section{Memory Operations and Sensory Inputs}
\label{sec:MM}

\subsection{Querying and Association}
\label{sec:sampling}

The models so far calculated the probability that a triple or quadruple is true.
For querying, we re-normalize the model and $\exp \beta \theta_{s, p, o}$  becomes proportional to the probability that the memory \textit{produces the triple} $(s, p, o)$ as a generative model. If the inverse temperature $\beta$ is high, only highly likely triples are produced.\footnote{For nonnegative models we can replace $\exp \beta \theta_{s, p, o}$ with  $\theta_{s, p, o}$.}\footnote{Note that in associative memories a similar concentration on the most likely interpretations can be achieved by  exponentiation~\cite{hintzman1984minerva}  or by using polynomial energy functions~\cite{krotov2016dense}.}
 Thus query answering can be implemented as  a  sampling process~\cite{LME-Nips2015}.
   The set of generated  triples then  represents  likely query answers. As an example,  by fixing  $s=\textit{Jack}$ and $p=\textit{likes}$, querying would generate likely entities as objects that \textit{Jack} is fond of.


  A recall {of a past previous events }at time $t'$  simply means that triples are generated from episodic memory using Equation~\ref{eq:epis}      with $\mathbf{a}_{e_{t'}}$ fixed  (see also Equation~\ref{eq:decode}). The pattern $\mathbf{a}_{e_{t'}}$ might have been generated by the activation of $e_{t'}$ or by some associative process.

  To recall what is known about an entity $i$, one applies  $\mathbf{a}_{e_{i}}$ and generates {triples} $(s=i, p, o)$ from semantic memory, describing facts about $i$.
  {Alternatively, one can generate quadruples $(s=i, p, o, t)$ from episodic memory, describing events that $i$ participated.} The set of triples could be used by the language modules (in the brain this would be  Broca's area) to generate language, i.e. to verbally describe an episodic or semantic  memory.
 It is also straightforward to generate an association, e.g., to retrieve entities semantically similar to, let's say,  $\mathbf{a}_{e_{i}}$.

\subsection{New Memories and Semantic Decoding of Sensory Inputs}


In some applications, information enters the memory system via sensory inputs, like vision and audition.
 The sensory input generates a latent representation
\begin{equation}\label{eq:sensMod}
 \mathbf{h}_t  = \mathbf{f}^M(\mathbf{u}_{t} )
\end{equation}
where $\mathbf{u}_{t}$ represents the sensory memory  (e.g., the image) at time $t$, and where the multivariate function $\mathbf{f}^M(\cdot)$ might be realized by a deep convolutional neural network (CNN).
If memorable, this latent representation is stored as $\mathbf{a}_{e_{t}} \leftarrow \mathbf{h}_t$ and bounded to the index $e_t$; it   can be semantically decoded by substituting  this $\mathbf{a}_{e_{t}} $ into  the episodic tensor model in Equation~\ref{eq:epis}. {Thus a memory trace or  engram consists of the pair $(e_t,  \mathbf{a}_{e_{t}})$.}

\section{A Cognitive Perspective}
\label{sec:cog}

\subsection{Forming Episodic Memories}
\label{sec:fem}

We relate our mathematical approach to the hippocampal memory indexing theory, which is one of the main theories for forming episodic memories~\cite{teyler1986hippocampal,teyler2007hippocampal} (Figure~\ref{fig:HMIT}). As in Equation~\ref{eq:sensMod}, a sensory input $\mathbf{u}_t$ activates a hierarchical multi-layered activation pattern $\mathbf{h}_t$ in the sensory layers of the cortex. As indicated in Figure~\ref{fig:HMIT},  this pattern is more specific towards the higher sensory layers. Thus, as also known from deep convolutional neural networks (CNNs), for a given sensory input, only few (but varying) nodes are activated on top of the hierarchy. Following the theory, an index  $e_t$ is formed in the hippocampus for sensory impressions important to be remembered by forming a map from the higher layers in the sensory hierarchy to the index. Since the representation of a sensory input becomes more sparse toward higher order sensor processing layers, the representations for different sensory inputs become also more orthogonal towards higher layers. In a sense, the index is simply the top of the hierarchy: the complete  index might involve a large number of neurons but a given sensory input only activates a small number of them, realizing a   sparse (pseudo-) orthogonal distributed representation. \footnote{In machine learning  terms,  $e_t$ is a new class trained by  one-shot learning  where the activation pattern $\mathbf{h}_t$ is reflected as a typically sparse weight pattern   $\mathbf{a}_{e_t}$.} A feature typically not present in CNNs is that the connection is assumed to be bidirectional and  the activation of index $e_t$ can reactivate the pattern $\mathbf{{h}}_t = \mathbf{a}_{e_t}$  in memory recall by back projection,  retrieving a sensory pattern of the past episode. {Of course, back propagation can only realise the gist of an episode.}

The index performs a binding of memory patterns. Instead of the hippocampus indexing
all of neocortex, it is probable that it participates in a hierarchical
indexing scheme whereby it indexes the association cortex (including the MTL (medial temporal lobe) and the
  ventral and dorsal streams),  which then indexes the rest of neocortex~\cite{teyler2007hippocampal}.  {Thus, the hippocampal area  sits at the end of a perceptual hierarchy, receiving input from lower levels and projecting back to them~\cite{moscovitch2016episodic}.}
An important biological function would be to recall previous episodic memories that are similar to $\mathbf{{h}}_t$ and to associate past emotional impressions, predictions (what happened next),  past actions, and past outcomes of those actions.

The formation of the index in the hippocampus is highly complex~\cite{teyler2007hippocampal,rolls2010computational} and one should interpret the index and the associated weight patterns {as functional descriptions,}  and not as biological implementations of indices or explicit  synaptic weights.
Details on the biological aspects of the index formation in the hippocampus can be found in the Appendix.

Evidence for time cells in the hippocampus (CA1) has  recently been found~\cite{eichenbaum2014time}.
In fact, it has been observed that the adult macaque monkey forms a few thousand new neurons daily~\cite{gluck2013learning}, possibly to encode new information. These time cells might be related to the {indices } $e_t$.

We can look at Equation~\ref{eq:sensMod} as an encoder for sensory inputs,  followed by a semantic decoder with
\begin{equation}\label{eq:decode}
  P(s, p, o | t) \propto \exp \beta  f^e(\mathbf{a}_{e_s}, \mathbf{a}_{e_p}, \mathbf{a}_{e_o}, \mathbf{a}_{e_{t}} = \mathbf{h}_t)
\end{equation}
which assumes the form of a generalized nonlinear model. Encoder-decoder networks are the basis for neural machine translation systems~\cite{cho2014learning}.  ~\cite{Baier2017ISWC} is an example where semantic decoding is applied to real-world images.

{In cognitive neuroscience, the memory recollection stages we have just described  sometimes are  referred to as internal and external stages~\cite{moscovitch2016episodic}. The internal cue generates the  pattern $\mathbf{a}_{e_{t'}}$  by the activation of $e_{t'}$ or by some associative process. It
  involves a rapid and unconscious interaction between the cue, which in turn reactivates the neocortical traces bound with it.
  In the second stage,  cortical processes operate on the output of the first stage to reinstate the conscious experience of the episode.
The involved  conscious experience is referred to as autonoetic consciousness,  a process that enables one to
relive episodes as  ``a subjective sense of time and of the
self as the one who experienced the episode and possesses the memory''~\cite{moscovitch2016episodic}. In our model, the second stage would be interpreted as the generation of triples from  episodic memory.}



\begin{figure}
	\centering
   \includegraphics[width=\columnwidth]{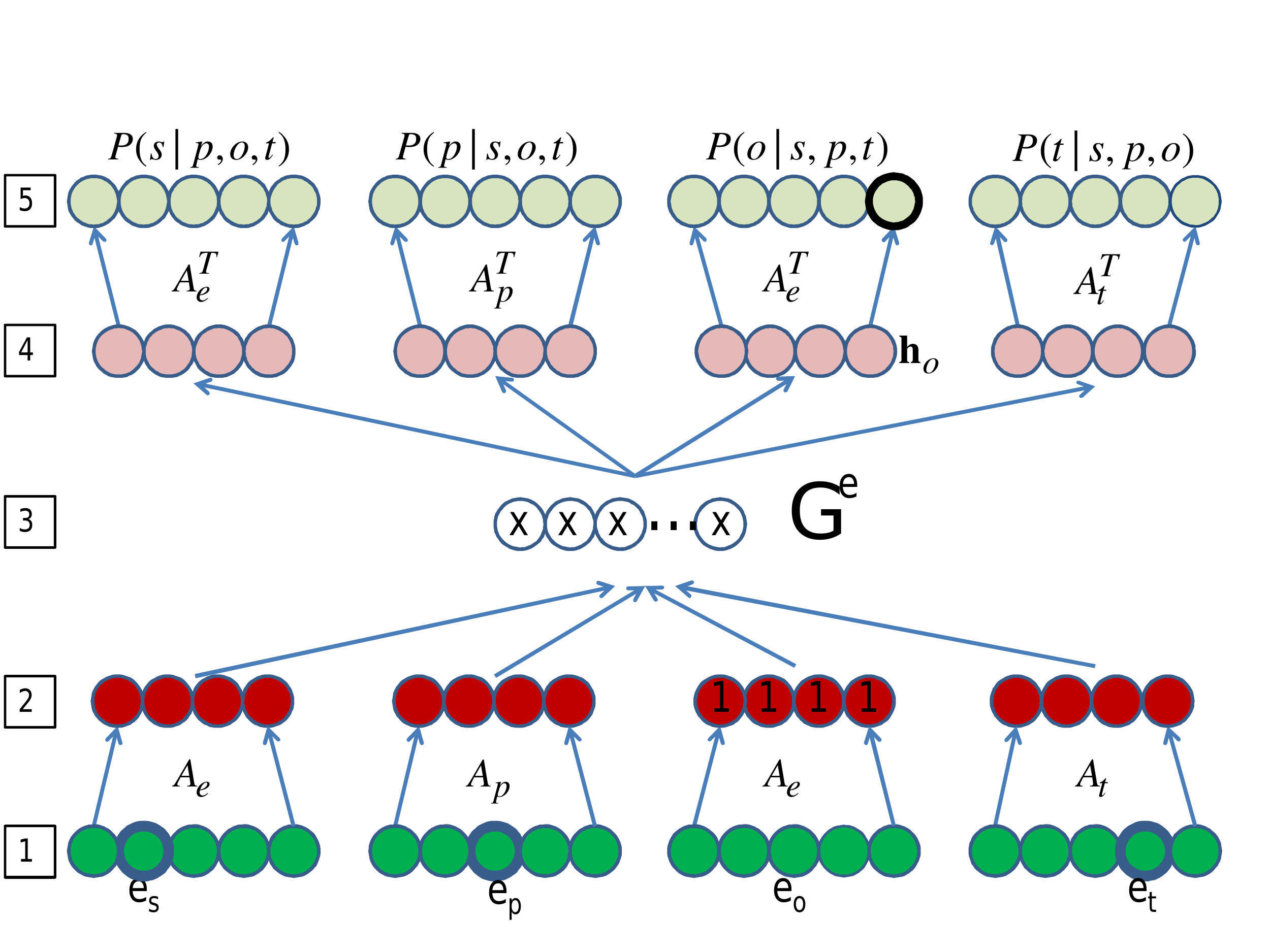}
	\caption{%
   An architecture implementing the episodic memory.
As concrete example, we assume that in the layer~1 the nodes for the corresponding  subject $s=2$ and predicate $p=3$ and time $t=4$ are activated. Thus the query would be: {Which objects are likely associated }with subject $s=2$ and predicate $p=3$ at the time $t=4$.
Layer~2 is the representation layer and the corresponding latent representations $\mathbf{a}_{e_{s}}$,  $\mathbf{a}_{e_{p}}$, $\mathbf{a}_{e_{t}}$ are represented.
Nodes in the product layer calculate
 ${a}_{e_s, r_1}
 \;   {a}_{e_p, r_2}
 \; {a}_{e_t, r_4}  \;
 g^e(r_1, r_2, r_3, r_4)$.
 Since in the example we query an object, the latent representation for the object is replaced by ones in layer 2.  In the second to right decoder of layer~4 we form
$ \mathbf{h}_o = \sum_{r_1=1}^{\tilde r} \sum_{r_2=1}^{\tilde r}
 \sum_{r_4 =1}^{\tilde r}
 {a}_{e_s, r_1}   \;   {a}_{e_p, r_2}     \; {a}_{e_t, r_4}  \;
 g^e(r_1, r_2, r_3, r_4)$. The layer 5 then samples an object from  $P(o|s,p,t) \propto \exp \beta \mathbf{a}_{e_o}^T \mathbf{h}_o$.  The columns of $A_e$, $A_p$, $A_t$ contain the latent representations of the entities, predicates and time steps.
By following different paths through the graph,  we can generate a sample from $P(s| p,  o, t)$,  from $P(p|s, o, t)$, or from $P(t| s, p,  o)$.
We can also sample from marginalized distributions. Since nonnegative  tensor models belong to the class of sum-product model~\cite{poon2011sum}, marginalization means to enter a vector of ones for quantities to be marginalized. For example to marginalize out $p$, we would enter a vector of ones $\mathbf{1}$ to the index layer for predicates in layer~1.
Biologically, this can be interpreted as a ``neutral'' input.
The figure shows nicely that  an entity only communicates with the rest of the network via its latent representation.  We assume that the processing steps in layers 1-5 are executed sequentially.
}
	\label{fig:teps}
\end{figure}

\subsection{Episodic and Semantic Memories}
\label{sec:ese}

At a certain abstraction level the brain has been modelled as  a graph  of computing nodes~\cite{bullmore2009complex}.

Here, we discuss distributed graphical implementations of the mathematical models. Figure~\ref{fig:teps} shows a graphical structure with five layers. In {this }example, \textit{s, p, t} are given and the goal is to find an \textit{o} that leads to likely {quadruples.} In the second
 layer the latent representations $a_{e_s}$, $a_{e_p}$, and $a_{e_t}$ are activated. The latent representation for the unknown object is set to ones as indicated. The third layer calculates
 ${a}_{e_s, r_1}
 \;   {a}_{e_p, r_2}
 \; {a}_{e_t, r_4}  \;
 g^e(r_1, r_2, r_3, r_4)$ and the fourth layer (second decoder from the right) calculates the sum
\[
\mathbf{h}_o = \sum_{r_1=1}^{\tilde r} \sum_{r_2=1}^{\tilde r}
 \sum_{r_4 =1}^{\tilde r}
 {a}_{e_s, r_1}   \;   {a}_{e_p, r_2}     \; {a}_{e_t, r_4}  \;
 g^e(r_1, r_2, r_3, r_4) .
 \]
 The fifth layer samples one or several objects based on
\begin{equation}\label{Eq:cond}
 P(o | s, p, t) \propto \exp \beta  \mathbf{a}_{e_o}^T \mathbf{h}_o .
\end{equation}
We exploit here that  a multi-linear expression as in Equation~\ref{eq:epis}, can be written as $\mathbf{a}_{e_o}^T \mathbf{h}_o$.\footnote{We can write the Tucker model as $f^e(\mathbf{a}_{e_s}, \mathbf{a}_{e_p}, \mathbf{a}_{e_o}, \mathbf{a}_{e_{t}}) = \mathcal{G}^e \bullet_1 \mathbf{a}_{e_s}  \bullet_2 \mathbf{a}_{e_p} \bullet_3 \mathbf{a}_{e_o} \bullet_4 \mathbf{a}_{e_t}$. $\mathcal{G}$  is the core tensor. $\bullet_n$ is the n-mode vector product~\cite{kolda_tensor_2009}. }
{ $\mathbf{h}_o$ can be interpreted as an activation pattern generated in neocortex, which is then matched with existing representations of entities $\mathbf{a}_{e_o}$. }

 By entering the vectors of ones at the latent representations in layer 2 for  $\mathbf{a}_{e_s}$, $\mathbf{a}_{e_p}$, or $\mathbf{a}_{e_t}$ and
 by taking different paths through the models, we   can  calculate the  conditional probabilities
 $P(s|p, o, t)$, $P(p|s, o, t)$, $P(t| s, p, o)$, respectively.
 With nonnegative tensor models, we can also marginalize: Since the (nonnegative) tensor models belong to the class of sum-product networks,  a marginalization simply means the application of all ones for the variable to be marginalized in the index layer 1.

 This permits us to, e.g., calculate $P(o|s, t)$, marginalizing out $p$.\footnote{Note that the final operation as described in Equation~\ref{Eq:cond} is the dot product of a vector generated  from the latent representations of $s, p, t$, i.e., $\mathbf{h}_o$,  with the latent representations of the object, i.e., $\mathbf{a}_{e_o}$. The general structure in Figure~\ref{fig:teps} is the same for different kinds of  tensor models, whose implementations vary in layer 5. For example, a RESCAL model~\cite{nickel_three-way_2011} calculates $\mathbf{h}_o = G^p \mathbf{a}_{e_s}$ which can be related to many classical associative memory models. The predicate matrix $G^p$ is a slice in RESCAL's core tensor.
Main differences to the classical associative memory models  are that here the factors are latent and that the system is trained end-to-end, whereas classical systems rely on some form of Hebbian learning~\cite{hopfield1982neural}. Also classical memory models are often auto-associative, i.e., their main concern is to restore a noisy pattern, whereas the models considered here are associative, in that they predict an object, given subject and predicate.}

 In ~\cite{tresp2017embedding}, we argue that a semantic memory can be generated from an episodic memory by marginalizing time.
 Thus, as an example,  to estimate $P(o|s)$ we would activate the index for $e_s$ in layer 1,
 apply vectors of ones at layer 1 for predicate  and time  (to marginalize out both) and apply a vector of ones at layer 2 for the object (the quantity we want to predict).
 The generation of semantic  memory from episodic  memory would biologically be very plausible.\footnote{This form of marginalization only works for nonnegative models.}
 { The marginalization of the time index can be implemented as a simple iterative process, as shown in Section~\ref{sec:conso}.}
  This relationship between both memories  is supported by cognitive studies on brain memory functions:  It has been argued that  semantic memory is information an individual has  encountered repeatedly, so often that the actual learning episodes are blurred~\cite{baddeley1974working,squire1987memory,conway2009episodic}. {Section~\ref{sec:conso} discusses consolidation and the relationship between episodic memory and semantic memory in more detail.}

\begin{figure}
	\centering
    \includegraphics[width=\columnwidth]{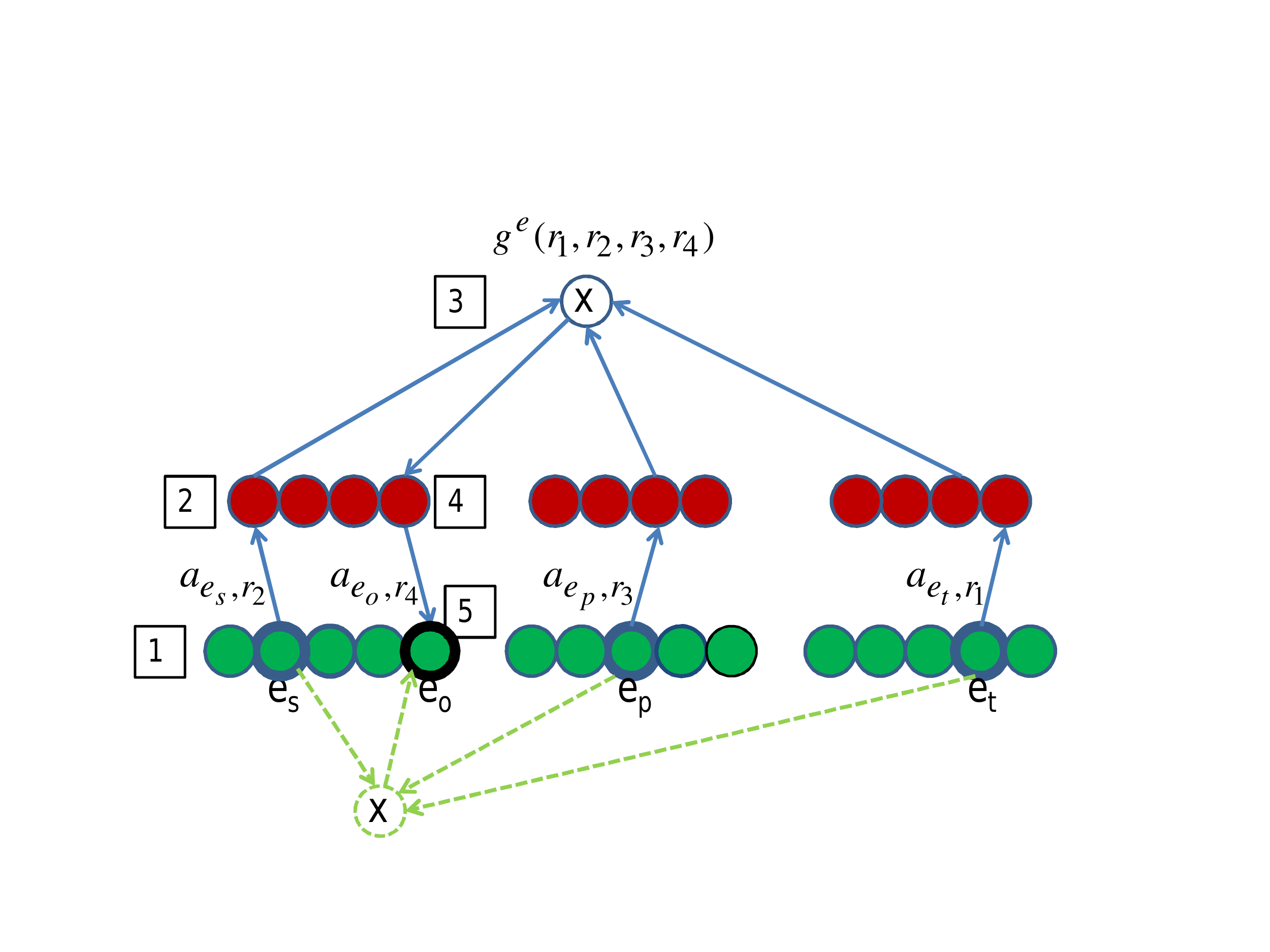}
	\caption{%
Here we show the information flow that involves a specific product node (one out of $\tilde r^4$ of such nodes) of the graphical representation in  Figure~\ref{fig:teps}.
The figure also shows that entities have unique representations (left), independent of their roles as subject or object. Also layers {2 and are 4 identical in Figure~\ref{fig:teps}, as are  layers 1 and 5, } and can be reduced to  single representations, as shown here. The nodes on the activation layer 2 are connected to the product node as shown which has a multiplicative  bias $g^e(r_1, r_2, r_3, r_4)$. As discussed in the caption of Figure~\ref{fig:teps},  the product node calculates  ${a}_{e_s, r_1}
 \;   {a}_{e_p, r_2}
 \; {a}_{e_t, r_4}  \;
 g^e(r_1, r_2, r_3, r_4)$. In layer 4 (which is identical to layer 2), the sum over all product nodes is calculated to form $\mathbf{h}_o$, and an  object is sampled from the resulting probabilities in layer 5 (which is identical to layer 1).
  {Although the complexity of the network is approximately $\mathcal{O}(\tilde r^4)$},
 this number is fixed and does not grow with the number of  entities, predicates or time steps.  In contrast, in an explicit representations of facts (lower part of the figure in pale green), which is the basis of other tensor-based memories~\cite{halford1998processing},
  the number of product nodes grows proportional to the number of episodic facts to be stored. This explicit fact  representation might be useful for ``unusual'' facts that cannot easily be represented by factorization models and might be related to classical semantic network memory models~\cite{collins1975spreading}, knowledge graphs~\cite{singhal_introducing_2012} and hypergraphs.
}
	\label{fig:path-teps}
\end{figure}

It appears from Figure~\ref{fig:teps} that some representations are redundant. For example $A_e$ appears several times in the figure. Also layers 2 and 4 both represent latent representations and layers 1 and 5 both represent indices. Figure~\ref{fig:path-teps} shows a computational path without redundancies. It also reflects that the entity sets for subject and object are really identical.

Tensor models have been used previously as memory models but the main focus was the learning of simple (auto-) associations~\cite{hintzman1984minerva,humphreys1989different,osth2015sources} and  compositional structures~\cite{smolensky1990tensor,pollack1990recursive,plate1997common,halford1998processing}. In the tensor product approach (Smolensky~\cite{smolensky1990tensor}),  encoding or binding is realized by a tensor product (generalized outer product), and composition by  tensor addition.  In the STAR model~\cite{halford1998processing}, predicates are represented as tensor products of the components of the predicates; in the triples considered here, these would be subject, predicate and object.  None of these approaches uses tensor \textit{decompositions}.

\subsection{Future Work: Perception as an Active Process}

Equation~\ref{eq:decode} describes a basic encoder-decoder system.  It can be made more powerful by integrating attention mechanisms, serial processing and recurrent structures~\cite{cho2014learning,cho2015describing,badrinarayanan2015segnet}. As part of future work we will explore recent developments in encoder-decoder approaches for sensory processing and semantic decoding.
 In~\cite{Baier2017ISWC} a semantic prior is combined with a triple extraction process that involves visual attention based on extracted bounding boxes.  This is another line of research worth exploring.

%

\section{Consolidation}
\label{sec:conso}

\subsection{Memory Consolidation}
\label{subsec:mc}

Memory consolidation is a controversial issue and there exist several theories concerning the system consolidation of memory. In the standard consolidation theory (SCT),  episodic memories from the hippocampal region, where memories are first encoded, are moved to the neo-cortex in a more permanent form of storage~\cite{squire1995retrograde}.   In this process the  hippocampus is ``teaching'' the cortex, eventually making episodic  memory hippocampus-independent: The neocortex then can support episodic memory indefinitely.

{An alternative view is given  by the multiple trace theory (MTT), which clearly distinguishes between episodic and semantic memory~\cite{nadel1997memory,nadel2000multiple,winocur2010memory}.  MTT argues that the hippocampus is always involved in the retrieval and storage of \textit{episodic} memories but that it is less  necessary during the encoding and use of \textit{semantic} memories. Episodic memory is never transferred from hippocampus to neocortex but episodic memories  are used to train a  semantic memory, located in neocortex (with prominent roles of the  ventromedial prefrontal cortex (vmPFC) and the  anterior temporal lobe~\cite{moscovitch2016episodic}). MTT is a basis for the  complementary learning systems (CLS) framework~\cite{mcclelland1995there,o2014complementary,kumaran2016learning}.  According to {that} theory, effective learning requires two complementary systems: one, located in the neocortex, serves as the basis for the gradual acquisition of structured knowledge about the environment, while the other, centered in the hippocampus, allows for rapid learning of the specifics of individual items and experiences.
{A central claim of the CLS framework is that \textit{the hippocampus encodes information in a
qualitatively different way than the neocortex}~\cite{o2014complementary}:  ``In other
words, there is not a literal ‘‘transfer’’ of information out of the hippocampus to the cortex, but rather the cortex learns its own, more distributed, version of what the hippocampus had
originally encoded.''}}


Our theory contains components of both, SCT and MTT,  and proposes some new perspectives.

\begin{description}
  \item[Role of Hippocampus:] In our theory, the main purpose of the hippocampus is  to form the indices $e_t$ and link it to the sensory processing hierarchy by $\mathbf{a}_{e_t}$. Thus an engram of memory trace is the pair $(e_t, \mathbf{a}_{e_t})$. This is accordance with both SCT and MTT. Of course the hippocampus also plays an active  role in consolidation and in other functions. But this is beyond the scope of this paper.

  \item[Semantic Decoding of $\mathbf{a}_{e_t}$:]  The semantic decoding of an episodic event would
   produce sets of triples describing $\mathbf{a}_{e_t}$.  Our episodic tensor model would require the representations for entities  $(e_i, \mathbf{a}_{e_i})$ and predicates $(e_p, \mathbf{a}_{e_p})$ as well. This decoding is not likely  happening in hippocampus, so semantic decoding of (non-consolidated) episodic memory involves both the hippocampus (for $(e_t, \mathbf{a}_{e_t})$ ) and the neocortex (for $(e_i, \mathbf{a}_{e_i})$, $(e_p, \mathbf{a}_{e_p})$ ).

     \item[Consolidation of Episodic Memory:]
 Implementing the consolidation of episodic memory into neocortex, as proposed by SCT, would be straightforward:  time indices { $e_t$ and their connection  patterns  $\mathbf{a}_{e_t}$  of memorable episodes  would also build  }  representations in neocortex.    A simple mechanism would be that the hippocampal indices reactivate past memory activation patterns which then trigger new index formations in neocortex by replay. Thus in this model, the representation of episodic memory would be very similar in hippocampus and in neocortex.  MTT  would  not agree with his hypothesis.

\end{description}




For the transfer of episodic memory into semantic memory, we propose several  mechanisms that would agree with our theory.   In our model this would be realized by a tensor decomposition model as in Equation~\ref{eq:sem}.
The semantic memory would have a graphical structure similar  to the episodic memory in Figure~\ref{fig:teps}. The time-related structures on the right would be removed and the core tensor of the episodic memory $\mathcal{G}^e$ would be replaced by the core tensor for the semantic memory $\mathcal{G}^s$.
Thus the challenge is to calculate the semantic core tensor $\mathcal{G}^s$.        We now discuss different options.

\begin{description}
  \item[Marginalization] The first one is that semantic memory is marginalized episodic memory, as described in Subsection~\ref{sec:ese} and proposed and studied in~\cite{tresp2017embedding,YunpuErik2018}.       $\mathcal{G}^s$ can be derived from $\mathcal{G}^e$ very efficiently, using  the iteration
   \[
      g^s(r_1, r_2, r_3) = g^s(r_1, r_2, r_3) +   \sum_{r_4} a_{e_{t}, r_4} g^e(r_1, r_2, r_3, r_4)  \; \;\;\; \; \;\;\;  t = 1, 2, \ldots    .
      \]
  This calculation is the marginalization step described earlier and might be implementable by some biologically plausible process.   {The iteration can be performed at the time of perception, i.e., the semantic decoding of sensory input.}
  This could realize one episodic training process, as required in MTT. In a concrete implementation, some form of dynamic normalization might be required. { Note that although the  core tensor of the  episodic memory  is fixed or only slowly adapting,  the core tensor of the semantic memory is constantly changing by integrating new information.  }

  \item[Explicit Triple Generation:]   Here  the episodic memory ``teaches'' the semantic memory by  explicitly generating  triples using Equation~\ref{eq:decode}, produced during perception or during replay. This mechanism would be in accordance  with MTT.

  \item[Explicit Triples Stored in a Knowledge Graph:]  The semantic tensor models discussed so far perform well in generalization to new facts and would represent what is sometimes  called the ``gist'' of memories, as discussed in~\cite{winocur2010memory,o2014complementary}.  In contrast,  semantic memory can also be quite sharp: ``Munich is part of Bavaria'' is a hard fact and also perceived as such.
      We propose an explicit semantic memory storage where a subset of the triples generated in the  episodic decoding is stored as an explicit semantic networks, forming knowledge graphs (see pale green graph in the bottom of Figure~\ref{fig:path-teps}). In the corresponding hypergraph,  indices for entities and for predicates are the nodes.  This representation cannot generalize to new facts. Such an explicit memory is not currently part of either SCT or MTT.
%
%
%

%
%
%
%
\end{description}






{In all three approaches,  episodic memory  teaches the neocortex to form a separate semantic memory, which is a core assumption in MTT and CLS.}
The semantic memory becomes a time-independent memory that does not rely on episodic memory, in particular not on $e_t$ and $\mathbf{a}_{e_t}$.

{In essence,  episodic memory is represented by $(e_t, \mathbf{a}_{e_t})$, and we would agree that this is different from the representation of semantic memory.   On the other hand, the semantic decoding of episodic memory   relies on entity representations $(e_i, \mathbf{a}_{e_i})$ and predicate representations $(e_p, \mathbf{a}_{e_p})$.
 In the marginalization approach, which we believe is the most interesting one, semantic memory is based on the same latent representations. }

Note that the forming of semantic memories might not be the sole role of replay but it might also be used to improve implicit memories for tasks like  prediction, control,  and reinforcement learning~\cite{hassabis2017neuroscience}.

{ In any case, consolidation is an involved process, likely happening during sleep and involving specific oscillation patterns such as sharp waves and ripples~\cite{squire1995retrograde,frankland2005organization,schwindel2011hippocampal,marshall2007contribution,stickgold2005sleep}.
The relevance of neural oscillations, in particular
 the coupling of theta and gamma rhythms, has been discussed in~\cite{nyhus2010functional}.
}

%
%
%


%
%


\subsection{Representations for Entities}

We would suggest that indices for entities $e_i$  are formed in MTL (the hippocampus and its surrounding cortex). In that area, \cite{quiroga2012concept} identified concept cells with  focussed responses to individuals like the actresses Jennifer Aniston and Halle Berry. The paper proposes that an assembly of concept cells would encode a given concept, in our case an index $e_i$.  Thus, the assumption we are making is that an index   might involve the activation of one or a small assembly of neurons~\cite{quiroga2013brain},  uniquely identifies an entity (see Subsection~\ref{sec:fem}).

%
%
%

%
%


Some recent studies have found detailed  semantic maps in neocortex~\cite{Huth2016}. At this point it is not clear if they might represent indices or components of the latent representations.
  The latent representations $\mathbf{a}_{e_i}$ associated with the indices would certainly be distributed in the brain. Thus the representation pattern for the concept ``hammer'' might activate brain regions associated with the typical appearance of a hammer (visual cortex), but also with the sound of hammering (auditory cortex) and with the activity of hammering (motor cortex)~\cite{kiefer2012conceptual}. As another example,  if  a subject recalls a person,  all sensory impressions of that person are restored. Note that each entity has a distributed representation since an index $e_i$  and its representation  $\mathbf{a}_{e_i}$ are always jointly activated. As Quiroga formulates~\cite{quiroga2013brain}: ``... the
activation of concept cells brings the particular concept into
awareness to embed it within its related circumstances and to
enable the creation of associations, memories and the flow of
consciousness. At the same time, the activation of concept cells
points towards to, and links, related and more detailed and semantic representations in different cortical areas. These concepts
are the subjective meaning we attribute to external stimuli, depending on how we may want to remember them.''


\section{Conclusions}
\label{sec:concl2}

The work presented in this paper  is an extension to the hippocampal memory
indexing theory, which  is one of the main detailed theories about the  forming of episodic memory. We have extended the model by also considering semantic decoding into  explicit triples and by providing explicit models for episodic and semantic memory. Our approach is built upon  latent representations of
generalized entities representing, e.g., objects, persons, locations, predicates, and time instances.  As in the hippocampal memory
indexing theory, an activation of an index for a past memory would activate its representation in association cortex which might reconstruct  sensory impressions of past memories. A very useful function of an episodic memory would be to recall what happened  after a past episode that is similar to the current situation and what action was applied and with what consequences. As a new contribution, we propose that past representations are also decoded semantically, producing triples of past memories. Semantic decoding might be an important intermediate step to generate language, i.e., to explicitly report about perceived sensory inputs, past episodes and semantic knowledge. Language, of course, is  a faculty specific to humans.

The activations patterns $\mathbf{a}$ might represent an entity,  a predicate, or a time instance on the subsymbolic level, whereas the corresponding index $e$ represents a concept on a discrete symbolic level.


We have discussed how the transfer of episodic from hippocampus to neocortex could be realized un our model, This transfer is at the core of the standard consolidation theory (SCT). MTT would not assume such a mechanism.
For the formation of semantic memory we have introduced a number of concepts. {In particular,  we show explicitly how episodic memory can teach the neocortex to form a semantic memory, which is a core issue in MTT and CLS.}
{ Although our theory would agree that semantic memory can be realized separately from episodic memory, we would propose that the semantic decoding, i.e. the declarative part of episodic memory,   is closely linked to semantic memory,  since both rely on the same latent representations of entities and predicates.}

%


We have demonstrated that  episodic and semantic memories can be modelled using tensor decompositions.  In a way  this is an existence proof, showing  that there are biologically plausible architectures that can implement episodic and semantic memory.  The brain might use different mathematical structures, although
 the marginalization of episodic memory,  which is the basis of one approach to  semantic memory discussed, is strictly only possible in the way described for sum-product models, such as   tensor decompositions.

In this paper, we have focussed on memory models. We propose that other functions like prediction, planning, reasoning and decision making would use the indices and their latent representations as well~\cite{esteban2016predicting,yang2016predictive}.
{In particular short-term memory might exploit  both semantically decoded indices and their latent representations.}
Cognitive control functions and working memory functions are typically associated with prefrontal cortex.

\subsection*{Acknowledgements}

Fruitful  and insightful discussions with Hinrich Schuetze and Maximilian Nickel are gratefully acknowledged.

\bibliographystyle{plain}
  \bibliography{TensorMemory}

\section*{Appendix: More Details  on Hippocampal Memory}



Evidence suggests that dentate gyrus (DG) and CA3 subregions of hippocampus sustain the sparsest active neurons. The DG subregion processes the incoming inputs prior to the associative learning in the CA3 region; it performs pattern separation and pattern orthogonalization in a sense that the overlapping of neurons firing patterns of similar inputs becomes reduced. It is biologically realized by the sparse firing activity of DG cells as well as sparse connections between DG and CA3 cells via mossy fiber. Sparse projection from DG to CA3 further reduces the degree of correlation inside CA3 and increases the storing capacity \cite{HIPO:HIPO450020209,cerasti2010informative}. The concept of an index  in our model is inspired by the sparse representation in the DG/CA3 subregions of the hippocampus.

It is predicted that the retrieval of hippocampus-dependent information from a retrieval clue depends on the CA3 subregion, which can be modeled as an attractor neural network with recurrent collateral connections. For each neuron the number of synapses connecting with the recurrent collaterals is supposed to be much smaller than the number of CA3 neurons;  this results in a diluted connectivity of the CA3 recurrent network. Diluted connectivity increases the number of attractor states (or stable states) of recurrent network and enhances the memory
capacity \cite{rolls2012advantages,rolls2006computational}.

We assume  that a time index $e_t$ is formed if the sensory perception is novel and emotionally significant. Considerable recent works have argued that the hippocampus is responsible for the temporal organization of memories \cite{eichenbaum2013memory,eichenbaum2014time};  it becomes activated when a sequence of serial events are being processed. To be explicit, the CA1 subregion of the hippocampus is involved in the sequential memory for temporal ordered events. The CA1 subregion has direct connections with the CA3 subregion as well as the medial entorhinal cortex. There is an interesting hypothesis that the  medial entorhinal cortex might supervise the formation of sequence memories since the grid cells in the medial entorhinal cortex might provide both spatial information and temporal information  required for sequence encoding~\cite{kraus2013hippocampal}. A direct connection from medial entorhinal cortex to CA1 specializes the firing pattern of neurons in CA1 at different time instances. This process is implemented by competitive networks, which are  essential for the item association operation in CA3~\cite{rolls2000memory}. In our model this process is abstracted as follows: entorhinal cortex organizes time indices  in a sequence, $e_t$, $e_{t-1}$, $\cdots$, they can further trigger indices detecting and binding in the CA3 subregion,  and outputs a series of time-ordered triples $(e_s,e_p,e_o)_t$ that together form a sequential episodic memory.

\end{document}